\documentclass[10pt,twocolumn,letterpaper]{article}

\usepackage[pagenumbers]{cvpr}      % To produce the REVIEW version
\usepackage{stfloats}
\usepackage{comment}

\definecolor{cvprblue}{rgb}{0.21,0.49,0.74}
\usepackage[pagebackref,breaklinks,colorlinks,allcolors=cvprblue]{hyperref}

\def\paperID{36} % *** Enter the Paper ID here
\def\confName{CVPR}
\def\confYear{2025}

\title{OCC-MLLM-CoT-Alpha: Towards Multi-stage Occlusion Recognition Based on Large Language Models via 3D-Aware Supervision and Chain-of-Thoughts Guidance}

\author{
Chaoyi Wang$^{1}$, Baoqing Li$^{1}$\thanks{Corresponding author}, Xinhan Di$^{2}$\footnotemark[1] \\
$^{1}$Shanghai Institute of Microsystem and Information Technology, CAS, China \quad
$^{2}$Giant Network, China \\
{\tt\small chaoyiwang@mail.sim.ac.cn, sinoiot@mail.sim.ac.cn, deepearthgo@gmail.com}
}
\begin{document}
\maketitle
\begin{abstract}
Comprehending occluded objects are not well studied in existing large-scale visual-language multi-modal models. Current state-of-the-art multi-modal large models struggles to provide satisfactory results in understanding occluded objects through universal visual encoders and supervised learning strategies. Therefore, we propose OCC-MLLM-CoT-Alpha, a multi-modal large vision language framework that integrates 3D-aware supervision and Chain-of-Thoughts guidance. Particularly, (1) we build a multi-modal large vision-language model framework which is consisted of a large multi-modal vision-language model and a 3D reconstruction expert model. (2) the corresponding multi-modal Chain-of-Thoughts is learned through a combination of supervised and reinforcement training strategies, allowing the multi-modal vision-language model to enhance the recognition ability with learned multi-modal chain-of-thoughts guidance. (3) A large-scale multi-modal chain-of-thoughts reasoning dataset, consisting of $110k$ samples of occluded objects held in hand, is built. In the evaluation, the proposed methods demonstrate decision score improvement of $15.75\%$,$15.30\%$,$16.98\%$,$14.62\%$, and $4.42\%$,$3.63\%$,$6.94\%$,$10.70\%$ for two settings of a variety of state-of-the-art models.
\end{abstract}    
\section{Introduction}
\label{sec:intro}
Recent advances in multi-modal large language models (MLLMs) like GPT-4o~\cite{openai2023gpt4} have significantly enhanced vision-language understanding. However, reasoning about occluded objects is not well explored, essential for various real-world applications ~\cite{alazeb2024remote,dai2024advanced,ahmed2024dynamic,wang2020robust}.

Occluded object reconstruction has emerged as an effective method for understanding partially visible objects in real-world environments. Existing approaches have employed implicit feature fusion through geometric reasoning~\cite{brahmbhatt2020contactpose,cao2021reconstructing}, physical realism techniques~\cite{pham2017hand,tzionas2016capturing}, and signed distance fields (SDFs) representations, such as IHOI~\cite{ye2022whats} and geometry-driven SDF (gSDF)\cite{chen2023gsdf}. Recently, MOHO\cite{zhang2024moholearningsingleviewhandheld} utilized multi-view occlusion-aware supervision. These methods show great potential for improving occluded object understanding in multi-modal large language models (MLLMs).

Despite these efforts, understanding occluded objects in MLLMs remains challenging. Recent advances demonstrate that language instruction and multi-modal Chain-of-Thought (CoT) reasoning methods~\cite{wei2022chain,yao2024tree,xu2025llavacotletvisionlanguage,cesista2024multimodalstructuredgenerationcvprs,qiao2024prismframeworkdecouplingassessing}, which decompose complex tasks by spliting the process into perception and reasoning stages, ~\cite{ni2024visualo1understandingambiguousinstructions,zellers2019recognition,yue2024mmmu}. Additionally, methods like OCC-MLLM\cite{qiu2024occmllm} and OCC-MLLM-Alpha\cite{yang2024occmllm} have integrated specialized 3D modules and dual visual encoders. However, multi-modal CoT methods combined with 3D modules remain underexplored. Therefore, we propose integrating multi-modal CoT reasoning into vision-language models to improve occluded object understanding and the corresponding self-reflective ability.

We propose OCC-MLLM-CoT-Alpha (Multi-stage \textbf{OCC}lusion Recognition with \textbf{MLLM} via 3D-aware supervision and \textbf{C}hain-\textbf{o}f-\textbf{T}houghts Guidance), a multi-stage, multi-modal framework designed to understand and initially reason about occluded objects through progressive steps and self-reflection. At the first stage, we pre-train a multi-modal vision-language model and also train a 3D expert reconstruction model. At the second stage, the designed multi-modal Chain-of-Thoughts is learned through supervised learning and preference Learning. Moreover, a large-scale multi-modal occluded objects reasoning dataset is created, containing over $110k$ samples along with corresponding multi-modal Chain-of-Thought (CoT) annotations. 

\section{Method}
\label{sec:method}
The training process is consisted of two stages for the recognition of occluded objects tasks. At the first stage, we pre-train a multi-modal vision-language model and also train a 3D expert reconstruction model. At the second stage, the multi-modal Chain-of-Thoughts is learned through supervised learning and reinforcement learning, enabling the multi-modal vision-language model to develop the ability for both step-by-step reasoning and self-reflection, aiming at enhancing the recognition of the occluded objects. 

\begin{figure}[htbp] % figure* 
    \centering
    \includegraphics[width=1.05\columnwidth]{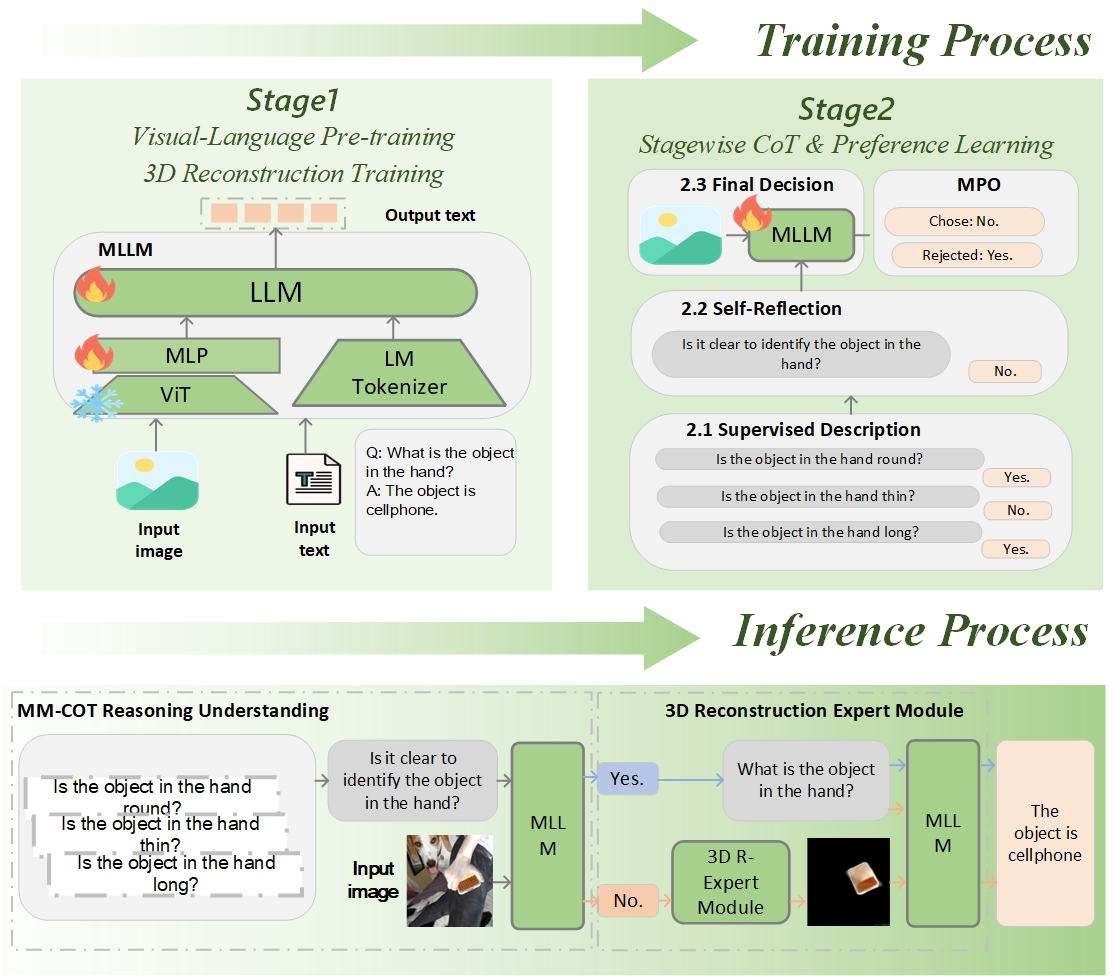}
    \caption{Step-by-Step Occlusion Reasoning Framework Using Multi-modal LLM with Stepwise Chain-of-Thoughts Guidance for Enhanced Object Recognition.}
    \label{fig:1}
\end{figure}

\subsection{Stage 1: Vision-Language Pre-training and 3D Expert Reconstruction Training}
\subsubsection{Vision-Language Pre-training}

The supervised Visual-Language learning training pipeline for a single model in our structure is organized into three stages, each aimed at enhancing the model's visual perception and multi-modal capabilities~\cite{chen2025expandingperformanceboundariesopensource}.

First, the process begins with MLP Warmup, where only the MLP projector is trained while both the vision encoder and language model remain frozen. Second, ViT Incremental Learning stage introduces training for both the vision encoder and the MLP projector. Third, the entire model comprising the vision encoder, MLP, and LLM is trained on high-quality multi-modal instruction datasets. The mechanism is represented as the following:
\begin{align}
&\min\limits_{\theta_{\mathrm{MLP}}, \theta_{\mathrm{LLM}}} \mathbb{E}{(\boldsymbol{x}_v, \boldsymbol{x}_t)}\sim \mathcal{D} \\
& \left[ \mathcal{L}_{LLM}( f_{\mathrm{LLM}}(f_{\mathrm{MLP}}(f_{\mathrm{ViT}}(\boldsymbol{x}_v; \theta_{\mathrm{ViT}}^*); \theta_{\mathrm{MLP}}); \theta_{\mathrm{LLM}}), \boldsymbol{y})  \right] \nonumber
\end{align}
where $\boldsymbol{x}_v$ is the visual input and $\boldsymbol{x}_v$ is the text input, $\boldsymbol{y}$ is the desired output, $\mathcal{D}$ is dataset. 
$\theta_{\mathrm{MLP}}$,$\theta_{\mathrm{LLM}}$ represents the trainable parameters of the MLP projector and LLM,
$\theta_{ViT}^*$ represents the frozen parameters of the vision encoder, $f_{\mathrm{ViT}}$, $f_{\mathrm{MLP}}$, and $f_{\mathrm{LLM}}$ represent the vision encoder, MLP projector, and language model functions respectively.

\subsubsection{3D Reconstruction Supervision Training}
We address hand-occlusion in single-view object reconstruction using a synthetic-to-real training strategy based on the MOHO model~\cite{zhang2024moholearningsingleviewhandheld}. The approach consists of two stages: (1) Synthetic Pre-training, using the SOMVideo dataset to handle occlusion via 3D Occlusion Handling (predicting complete object shapes from occluded views) and 2D Occlusion Awareness (predicting probabilistic hand coverage maps using an auxiliary head $\Gamma$); and (2) Real-world Fine-tuning on actual hand-object videos, described formally as follows:
\begin{align}
\min\limits_{\theta_{\mathrm{3D}}, \theta_{\Gamma}} &\mathbb{E}{(\boldsymbol{I}, \boldsymbol{S}_o, \boldsymbol{I}_{\mathrm{novel}})}\sim \mathcal{D} \\
[ &\mathcal{L}_{3D}(f_\mathrm{3D}(\boldsymbol{I} \odot \boldsymbol{S}_o; \theta_{\mathrm{3D}}) , \boldsymbol{I}_{\mathrm{novel}})] \nonumber\\
+ \lambda & [\mathcal{L}_{2D}(\Gamma(\boldsymbol{I} \odot \boldsymbol{S}_o; \theta_{\Gamma}) , \boldsymbol{M}_\mathrm{3D}) ] \nonumber
\end{align}
where $\boldsymbol{I}$ is the input image,
$S_o$is the occlusion mask,
$\boldsymbol{I}_{\mathrm{novel}}$ represents novel view supervisions,
$\textbf{M}_\mathrm{hand}$ is the ground truth hand coverage map,
$\odot$ represents element-wise multiplication,
$\lambda$ is a weighting factor for the 2D supervision loss.
$\theta_{\mathrm{3D}}$ and $\theta_{\Gamma}$ are the trainable parameters of the main model and auxiliary head respectively.
\subsection{Stage 2: Stagewise CoT and Preference Learning}
\subsubsection{Stage 2.1: Supervised Stage}
The Stagewise CoT process a step-by-step approach to object understanding through progressive self-questioning, which employs a structured series of fundamental attribute queries that guide the model toward more reliable object recognition. This supervised stage can be further divided into 3 sub-processes: 
\begin{align}
a_{\text{FD}} &= \mathcal{F}_{\text{CoT}}(\boldsymbol{x}_v) \\
&= f_{\text{CoT}}^{\text{FD}}(\boldsymbol{x}_v, \boldsymbol{x}_{t_{\text{FD}}}, f_{\text{CoT}}^{\text{SR}}(\boldsymbol{x}_v, \boldsymbol{x}_{t_{\text{SR}}}, f_{\text{CoT}}^{\text{SD}}(\boldsymbol{x}_v, \mathcal{X}_{\text{SD}})), \mathcal{R}_{\text{3D}}) \nonumber
\end{align}
This cascaded formula shows how information flows through our framework:
the Supervised Description (SD) stage takes the visual input $\boldsymbol{x}_v$ and predefined questions $\mathcal{X}_{\text{SD}}$ to produce answers $\mathcal{A}_{\text{SD}}$
These answers, along with the visual input $\boldsymbol{x}_v$ and self-reflection prompt $\boldsymbol{x}_{t_{\text{SR}}}$, feed into the Self-Reflection (SR) stage to produce a reflection result $a_{\text{SR}}$
Finally, the Final Decision (FD) stage combines all previous information with the 3D reconstruction model $\mathcal{R}_{\text{3D}}$ to produce the final decision $a_{\text{FD}}$.
%%%%%%%%%%%%%%%%%%%%%%%%%%%%%%%%%%%%%%%%%%%%%%%%
In summary, for the supervised stage, our overall goal is:

\begin{align}
&\min\limits_{\theta_\text{SD}, \theta_\text{SR}, \theta_\text{FD}} \mathbb{E}{(\boldsymbol{x}_v},\mathcal{X}_{\text{SD}}, \boldsymbol{x}_{t\text{SR}}, \boldsymbol{x}_{t\text{FD}}, \mathcal{R}_\text{3D})~\sim \mathcal{D} \\
& \left[ \sum\limits_{i \in {r,l,t}} \alpha_i \mathcal{L}_{\text{SD}}( f_i(\boldsymbol{x}_v, \boldsymbol{x}_{t_i}; \theta_{i}), \boldsymbol{y}_i) \right] \nonumber \\
& + \left[\lambda_\text{SR}\mathcal{L}_{\text{SR}}( f_\text{SR}(\boldsymbol{x}_v, \boldsymbol{x}_{t_\text{SR}}, \mathcal{A}_\text{SD}; \theta_\text{SR}), \boldsymbol{y}_\text{SR}) \nonumber \right]\\
& + \left[\lambda_\text{FD}\mathcal{L}_{\text{FD}}( f_\text{FD}(\boldsymbol{x}_v, \boldsymbol{x}_{t_\text{FD}}, a_\text{SR}, \mathcal{R}_\text{3D}; \theta_\text{FD}), \boldsymbol{y}_\text{FD}) \right] \nonumber
\end{align}
where $\mathcal{L}$ are the loss functions,  $\mathcal{D}$ represents the dataset, $f_r,f_l,f_t$ are the functions evaluating roundness, length, and thickness respectively, $\alpha_i$ are the weighting factors for each geometric description, $\boldsymbol{y}_{\text{i}}$ is the ground truth label for each question, $\lambda_{\text{SR}}$ and $\lambda_{\text{FD}}$ are weighting factors for self-reflection and final decision, $\boldsymbol{y}_{\text{SR}}$ is the label for self reflection and $\boldsymbol{y}_{\text{FD}}$ is the label for final decision.
\subsubsection{Stage 2.2 Mixed Preference Optimization}
In this stage, we apply Mixed Preference Optimization (MPO)\cite{wang2024enhancingreasoningabilitymultimodal}, combining three objectives into a balanced loss function. It integrates DPO\cite{rafailov2024directpreferenceoptimizationlanguage} for preference modeling without explicit rewards, BCO~\cite{jung2024binary} for absolute response quality assessment, and SFT (Supervised Fine-Tuning) as a generation objective~\cite{chen2024internvl,chen2024far,wang2024mpo}:
\begin{align}
    \mathcal{L}_{\text{MPO}} = w_{\text{p}}\mathcal{L}_{\text{p}} + w_{\text{q}}\mathcal{L}_{\text{q}} + w_{\text{g}}\mathcal{L}_{\text{g}} 
\end{align} 
where $\mathcal{L}_{\text{p}}$ is the preference loss, $\mathcal{L}_{\text{p}}$ is the quality loss, $\mathcal{L}_{\text{p}}$ is the generation loss.\\
The preference loss is defined as:
\begin{align}
    \mathcal{L}_{p} = \left[ \log \sigma \left( \beta \log \frac{\pi_\theta (\boldsymbol{y}_c \mid \boldsymbol{x})}{\pi_{\text{ref}} (\boldsymbol{y}_c \mid \boldsymbol{x})} - \beta \log \frac{\pi_\theta (\boldsymbol{y}_r \mid \boldsymbol{x})}{\pi_{\text{ref}} (\boldsymbol{y}_r \mid \boldsymbol{x})} \right) \right]
\end{align}
where $\beta$ is the KL penalty coefficient, and $\boldsymbol{x},\boldsymbol{y}_c, \boldsymbol{y}_r$ are user query, chosen response and rejected response respectively. The policy model $\pi_{\theta}$ is initialized from the reference policy model $\pi_\text{ref}$.\\
The quality loss is defined as:
\begin{align}
    \mathcal{L}_{\text{q}} &= \mathcal{L}_{\text{q}}^{+} + \mathcal{L}_{\text{q}}^{-},\\
    \mathcal{L}_{\text{q}}^{+} &= - \log \sigma \left(  \beta \frac{\pi_\theta (\boldsymbol{y}_c \mid \boldsymbol{x})}{\pi_{\text{ref}} (\boldsymbol{y}_c \mid \boldsymbol{x})} - \delta  \right),\\
     \mathcal{L}_{\text{q}}^{-} &= - \log \sigma \left( - \left( \beta \frac{\pi_\theta (\boldsymbol{y}_r \mid \boldsymbol{x})}{\pi_{\text{ref}} (\boldsymbol{y}_r \mid \boldsymbol{x})} - \delta \right) \right)
\end{align}
where $\mathcal{L}_{\text{q}}^{+}, \mathcal{L}_{\text{q}}^{-}$ represent the loss for chosen and rejected responses and $\delta$ represents the reward shift.\\
The generation loss is definded as:
\begin{align}
    \mathcal{L}_{g} = -\frac{\log \pi_{\theta}(\boldsymbol{y}_{c}|x)}{|\boldsymbol{y}_{c}|}
\end{align}
\begin{figure*}[!htbp] % figure* 
    \centering
    \includegraphics[width=0.76\textwidth]{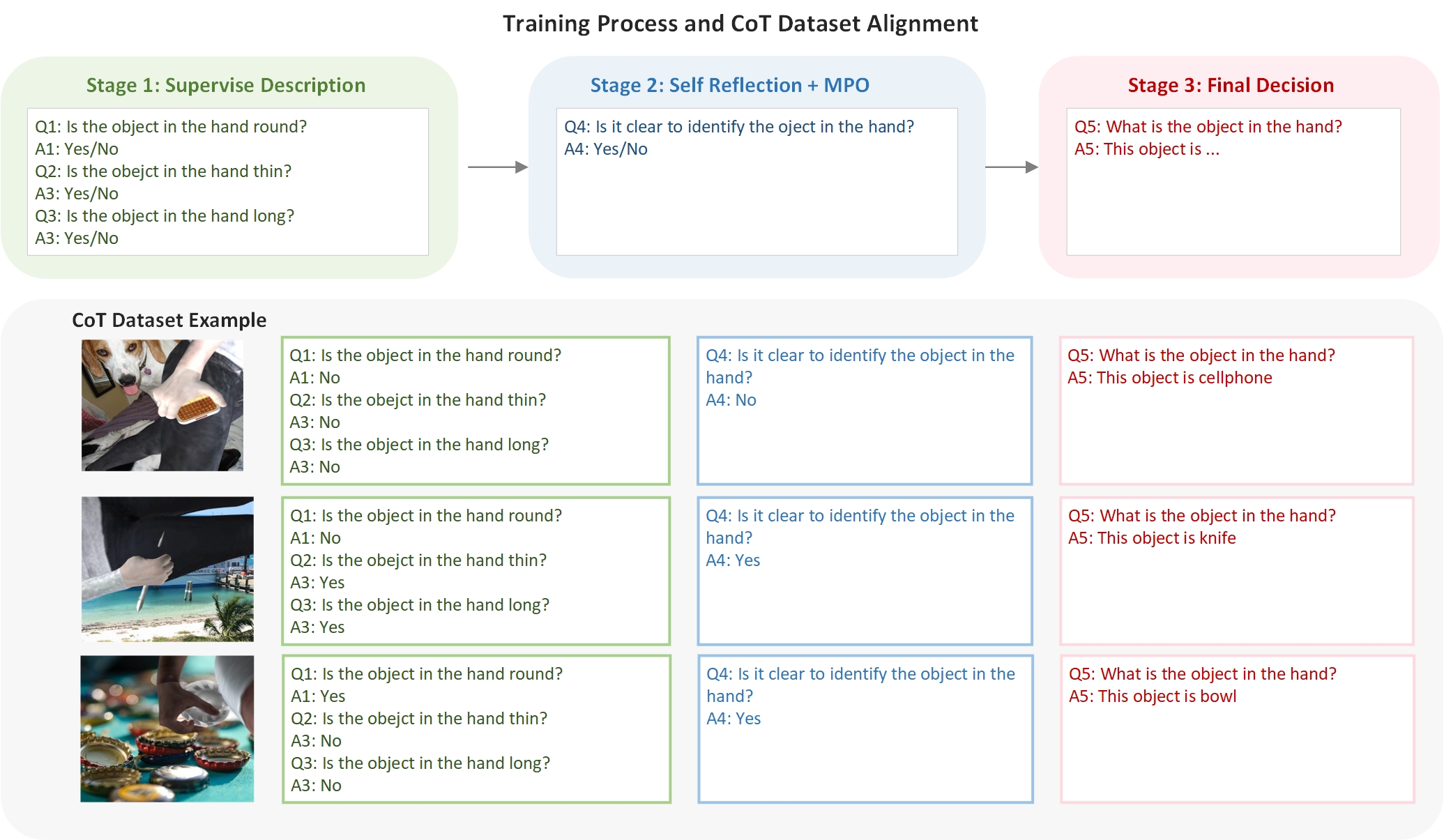}
    \caption{Step-by-Step Occlusion Reasoning Examples: Showcasing the Internal Chain-of-Thought Process.}
    \label{fig:2}
\end{figure*}
\section{Experiments}
\label{sec:experiments}
\subsection{Dataset}
The MLLM used in our experiment is pre-trained on a wide range of datasets, including single-image, multi-image, video, and text datasets, to handle multimodal tasks effectively. It incorporates diverse sources such as captioning datasets, general QA datasets, mathematics datasets, OCR datasets, and others~\cite{chen2025expandingperformanceboundariesopensource,chen2024allava,tong2024cambrian,liu2024mminstruct,schuhmann2022laion,maaz2024videogpt,li2024mvbench,jia2022egotaskqa,shah2019kvqa,dai2024multimodal,kuznetsova2020open,zhang2023llavar,shao2019objects365}. Similarly, the 3D reconstruction model is pre-trained using a vareity of datasets ~\cite{zhang2024moholearningsingleviewhandheld,hampali2020honnotate,chao2021dexycb}.

%synthetic-to-real framework~\cite{zhang2024moholearningsingleviewhandheld}. First, the model uses the SOMVideo dataset, a large-scale synthetic dataset containing 141,550 hand-object interaction scenes. Then the model is fine-tuned on real-world datasets including HO3D~\cite{hampali2020honnotate} and DexYCB~\cite{chao2021dexycb}, which feature real-world hand-object interaction videos.%SOMVideo provides occlusion-free multi-view supervision to address hand-induced occlusions during training. Then the model is fine-tuned on real-world datasets including HO3D~\cite{hampali2020honnotate} and DexYCB~\cite{chao2021dexycb}, which feature real-world hand-object interaction videos

We construct a multi-modal, chain-of-thought (CoT)-style dataset of $104,671$ image-text pairs for occluded object detection, based on the ObMan dataset~\cite{hasson19_obman}, which contains synthetic images of hands grasping objects. Our annotations introduce a structured reasoning process in three stages with five steps (see Fig.~\ref{fig:2}): (1) Description Stage: step-by-step questions on object attributes (e.g., round, thin, long); (2) Self-Reflection Stage: assesses clarity in object identification; and (3) Final Decision Stage: identifies the object explicitly (e.g., a cell phone).

\subsection{Evaluation}
We evaluate our model using three metrics: Description Score, Reflection Score, and Decision Score. The Description Score measures basic object recognition accuracy (``What is the object in the hand?"), reflecting fundamental visual understanding. The Reflection Score assesses the model’s judgment on visual clarity (``Is it clear to identify the object?"), deciding when to invoke a 3D Expert Reconstruction Model. Lastly, the Decision Score evaluates final identification accuracy, integrating Multi-Modal CoT reasoning with selective 3D reconstruction to enhance clarity for challenging cases before the final object identification.

\section{Results}

\begin{table*}
  \centering
  \begin{tabular}{l|ccc|ccc}
    \toprule
    Model & Description & Reflection & Decision & Description & Reflection & Decision\\
    \midrule
    \multicolumn{7}{l}{\textbf{Zero-shot}} \\
    \midrule
    GPT4v (Zero-shot)~\cite{openai2023chatgpt} & 0.0361 & - & 0.0361 & - & - & -\\
    GPT4o (Zero-shot)~\cite{openai2023chatgpt} & 0.1306 & - & 0.1306 & - & - & -\\
    \midrule
   \multicolumn{1}{l|}{\textbf{Setting}} & \multicolumn{3}{c|}{10K-Learning} & \multicolumn{3}{c}{100K-Learning} \\
    \midrule
    GPT4o (Learning)~\cite{openai2023chatgpt} & - & - & 0.5532 & - & - & - \\
    \midrule
    \multicolumn{7}{l}{\textit{MLLM-Qwen2-1B-Base}} \\
    Base(Learning) & - & - & 0.5500 & - & - & 0.6119 \\
    OCC-MLLM-CoT-Alpha & 0.6107 & 0.6624 & 0.6366 & 0.6155 & 0.6695 & 0.6390 \\
    \midrule
    \multicolumn{7}{l}{\textit{MLLM-Internlm2-2B-Base}} \\
    Base(Learning) & - & - & 0.5524 & - & - & 0.6189 \\
    OCC-MLLM-CoT-Alpha & 0.6119 & 0.6632 & 0.6369 & 0.6205 & 0.6766 & 0.6414\\
    \midrule
    \multicolumn{7}{l}{\textit{MLLM-Phi3-4B-Base}} \\
    Base(Learning) & - & - & 0.5571 & - & - & 0.6213 \\
    OCC-MLLM-CoT-Alpha & 0.6189 & 0.6958 & 0.6517 & 0.6223 & 0.72227 & 0.6644\\
    \midrule
    \multicolumn{7}{l}{\textit{MLLM-Internlm2.5-8B-Base}} \\
    Base(Learning) & - & - & 0.5751 & - & - & 0.6412 \\
    OCC-MLLM-CoT-Alpha & 0.6387 & 0.7085 & 0.6592 & 0.6785 & 0.7239 & 0.7098\\
    \bottomrule
  \end{tabular}
  \caption{Performance comparison across different models and training settings. For fine-tuning GPT-4o, we prepared 110,000 images, but 90,860 were automatically skipped due to the training policies, leaving 10,140 images for fine-tuning.}
  \label{tab:1}
\end{table*}

Table~\ref{tab:1} presents the performance comparison across different models and training settings. In zero-shot scenarios, GPT4o (0.1306) substantially outperformed GPT4v (0.0361). With fine-tuning, GPT4o's performance improved to 0.5532. Our OCC-MLLM-CoT approach demonstrated consistent improvements across all model variants. For 10K-learning, MLLM models showed progressive improvements with increasing model size, from Qwen2-1B (0.6366) to Internlm2.5-8B (0.6592). This trend continued in the 100K-learning setting, where Internlm2.5-8B achieved the highest performance with a Decision score of 0.7098, showing that both model capacity and training data significantly impact performance.
\section{Conclusion}
\label{sec:conclusion}
These results clearly demonstrate the effectiveness of our proposed OCC-MLLM-CoT-Alpha framework. In future work, we aim to: Improve the model's reasoning ability by introducing self-correction reinforcement learning; Design a more effective CoT process to enhance large model performance; Evaluate our approach on additional MLLM models and diverse datasets.

{
    \small
    \bibliographystyle{ieeenat_fullname}
    \bibliography{main}
}


\end{document}